\documentclass{bmvc2k}

\usepackage[T1]{fontenc}
\usepackage{epstopdf}
\usepackage{varwidth}
\usepackage{array}
\newcolumntype{x}[1]{>{\centering\arraybackslash\hspace{0pt}}p{#1}}

\title{GLAMI-1M:\\A Multilingual Image-Text Fashion Dataset}
\addauthor{Václav Košař}{vaclav.kosar@glami.cz}{1}
\addauthor{Antonín Hoskovec}{antonin.hoskovec@glami.cz}{1,3}
\addauthor{Milan Šulc}{milan.sulc@rossum.ai}{2}
\addauthor{Radek Bartyzal}{radek.bartyzal@glami.cz}{1}

\addinstitution{
 GLAMI.cz\\
 Křižíkova 148/34, 186 00 Prague 8,\\
 Czech Republic
}
\addinstitution{
Rossum.ai\\ Křižíkova 148/34, 186 00 Prague 8,\\
Czech Republic
}
\addinstitution{
FNSPE, Czech Technical University in Prague,\\
Břehová 7, 119 15, Prague 1,\\
Czech Republic
}

\runninghead{Košař, Hoskovec, Šulc, Bartyzal}{GLAMI-1M}

\begin{document}
%
\newcolumntype{M}{>{\begin{varwidth}{3.5cm}}l<{\end{varwidth}}}

%
%

%

%
%

\newgeometry{twoside,headsep=3mm,papersize={410pt,620pt},inner=15mm,outer=6mm,top=3mm,includehead,bottom=1mm,heightrounded}

\maketitle              
\begin{abstract}
We introduce GLAMI-1M: the largest multilingual image-text classification dataset and benchmark. The dataset contains images of fashion products with item descriptions, each in 1 of 13 languages. Categorization into 191 classes has high-quality annotations: all 100k images in the test set and 75\% of the 1M training set were human-labeled. The paper presents baselines for image-text classification showing that the~dataset presents a challenging fine-grained classification problem: The best scoring EmbraceNet model using both visual and textual features achieves 69.7\% accuracy. Experiments with a modified Imagen model show the dataset is also suitable for image generation conditioned on text.
The dataset, source code and model checkpoints are published at: \url{https://github.com/glami/glami-1m}. 
\end{abstract}


\begin{figure}[h!]
    \centering
    \vspace*{-0.6cm}
    \setlength{\tabcolsep}{1pt}
\renewcommand{\arraystretch}{0.8}
\small
\begin{tabular}{x{3cm} x{3cm} x{3cm} x{3cm}}
         \includegraphics[height=3.0cm]{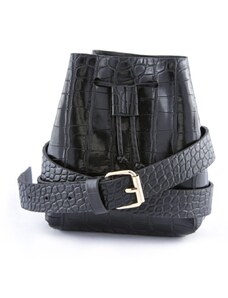}
 &  \includegraphics[height=3.0cm]{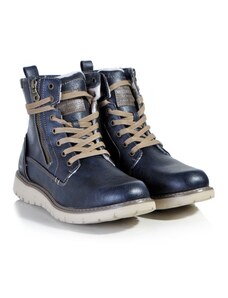}
 & \includegraphics[height=3.0cm]{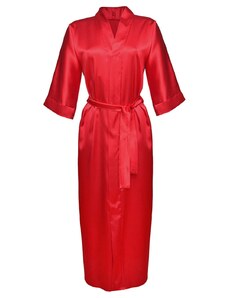}
 & \includegraphics[height=3.0cm]{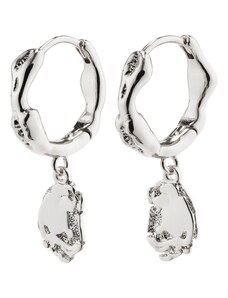} \\
 ANKA KEMER Kadın Heybe Çantalı Kemer 16x14 cm
 & Pánská kotníková obuv Mustang 4107-605-820 modrá
 & Ženski kopalni plašč DKaren Basic & Pilgrim Auskarai 'THANKFUL' sidabrinė \\
 (Turkey) & (Czechia) & (Slovenia) & (Lithuania) \\
 \textit{`womens-belts`} & \textit{`mens-boots`} & \textit{`womens-bathrobes`} & \textit{`womens-earrings`} \\

\end{tabular}
\caption{Examples from GLAMI-1M with their \textit{image}, \textit{name}, \textit{country} and \textit{class}. Available information not displayed: \textit{description}, \textit{label source} and \textit{item-ID}.}
\label{fig:exampleFront}
\end{figure}
\vspace*{-.5cm}

\restoregeometry

\section{Introduction}
Public datasets are a cornerstone of machine learning research: Cross-evaluation of different methods is possible thanks to public benchmarks with pre-defined training and test data splits. 
Pushing the envelope in machine learning often relies on considerable amount of training samples. For example, while the existence of Convolutional Neural Networks dates back to the 1980s \cite{fukushima1980neocognitron,lecun1989backpropagation}, the deep learning era in computer vision started with the success \cite{krizhevsky2012imagenet} on the ILSVRC 2012 challenge dataset \cite{ILSVRC15} commonly addressed as ImageNet. 
At the time of writing this paper, the best results reported\footnote{\url{https://paperswithcode.com/sota/image-classification-on-imagenet}} on ImageNet were achieved by an image-text model CoCa \cite{coca}, pre-trained on proprietary large-scale datasets JFT-3B \cite{jft3b} and ALIGN \cite{align} to produce joint image-text representation. Similarly, CMA-CLIP \cite{Liu2021CMACLIPCA} incorporated CLIP \cite{clip}, an ALIGN model \cite{align} predecessor, trained on proprietary WebImageText to achieve state-of-the-art image-text classification results on Fashion-Gen \cite{fashiongen}. These results suggest that image-text models have a great potential to aid image-based classification. 

Owing to the success of multilingual models \cite{bert,xml} and multimodal models \cite{clip,align}, datasets combining both multilingual and multimodal features are increasingly relevant to machine learning research (see \autoref{tab:multilingual}). 
However, public large scale image-text classification datasets \cite{recipe1m+,fashiongen,upmcFood101,Xie2019VisualEA} are still of rather limited size and language diversity (see \autoref{tab:imageTextClassification}). Note that, Recipe1M+ is not human annotated, rather its categories are extracted from recipe titles using statistical methods. In particular within the fashion domain, to the best of our knowledge, there is no large diverse multilingual text and image dataset (see \autoref{tab:fashion}) and machine translation cannot replace human produced text (yet).

In this paper, we introduce GLAMI-1M: the largest multilingual image-text classification dataset and benchmark. The dataset contains images of fashion products with item descriptions from an e-commerce platform. GLAMI-1M is a collection of 1.11M records representing a fashion product with an image, a name and description in one of 13~languages and a category within the GLAMI fashion search engine\footnote{at the point of extraction in 2022.}. 
Categorization into 191 classes has high-quality annotations: all 100k images in the test set and 75\% of the 1M training set were human-labeled. 

Organizing products from public listings into categories is an important problem in e-commerce platforms. 
Data from 
online production systems pose several challenges: dealing with imbalanced long-tailed class distributions \cite{shopify}, prior shift \cite{priorShift,vsipka2022hitchhiker}, noisy labels in case of rule-based annotations \cite{Sun2014ChimeraLC,distillingFromNoise} (as opposed to human labels), multimodal inputs \cite{shopify,foodi}, multilingual text \cite{shopify,foodi}, and utilizing available metadata \cite{df20}.


Datasets for related tasks and domains are reviewed in \autoref{sec:relatedWork}. The GLAMI-1M dataset and benchmark is introduced in \autoref{sec:GLAMI-1M}, including detailed analysis of its content and description of its creation. Baseline methods for image-text classification and text-conditional image generation are introduced in \autoref{sec:experiments}.  Additional details about the dataset and the experiments, and baselines for machine translation are provided in the supplementary material.

\section{Related Work}
\label{sec:relatedWork}

Large-scale image and multilingual text datasets are listed in \autoref{tab:multilingual}.
GLAMI-1M is the largest multilingual dataset for image-text classification.
Larger image-text datasets LAION-5B \cite{LAION5B}, WIT \cite{wit}, FooDI-ML \cite{foodi} are used for image-text retrieval, and miss standardized class labels.
Note that in \autoref{tab:multilingual}, we do not list translations of MS-COCO \cite{coco} such as  \cite{japaneseCoco,italianCoco,germanCoco,dutchCoco,chineseCoco,vietnameseCoco} as they are distributed in bilingual form, which does not pass the table's minimum of 3 languages. 

\begin{table}[htb]
\small
\centering
\caption{\label{tab:multilingual} Publicly available multilingual image-text datasets. 
Datasets with <3 languages and with <10k images or texts are omitted. The column task gives the most relevant task.}
\vspace{1mm}
    \setlength{\tabcolsep}{4pt}
\renewcommand{\arraystretch}{1}
\begin{tabular}{|l|l|l|l|l|m{2.9cm}|}
\hline
Dataset                          & Images        & Texts         & Langs     & Domain           & Task \\
\hline
LAION-5B \cite{LAION5B}          & 5.85B          & 5.85B         & 100+     & Web images        & image-text retr. \\
YFCC100M \cite{YFCC100M}         & 100M           & 100M          & 172      & Web images        & image-text retr. \\
WIT \cite{wit}                   & 11.5M          & 37.6M         & 108      & Wiki images       & image-text retr. \\
FooDI-ML \cite{foodi}            & 1.5M           & 9.5M          & 33       & Food, groceries   & text-image retr. \\
GLAMI-1M  & 968k          & 1.01M         & 13       & Fashion           & classification   \\
MultiSub (I4) \cite{multisub}    & 45k            & 180k         & 4        & subtitles, nouns  & fill-in-the-blank \\
Multi30k \cite{WMT18,multi30kGe,multi30kFr,multi30kCz}  & 30k    & 4 x 30k  & 4       & General           & machine translation \\
\hline
\end{tabular}
\end{table}

\begin{table}
\centering
\small
\caption{\label{tab:imageTextClassification} Publicly available image-text classification datasets.
Datasets with <30k images or texts are omitted.
}
\vspace{1mm}
\begin{tabular}{|l|l|l|l|l|l|l|}
\hline
Dataset                          & Images        & Texts         & Langs      & Domain           & Class. task    & Classes \\
\hline
Recipe1M+ \cite{recipe1m+}      & 13M           & 1M             & 1          & Recipes          & single-label    & 1047  \\
GLAMI-1M                        & 968k          & 1.01M          & 13         & Fashion          & single-label   & 191   \\
FashionGen \cite{fashiongen}     &325k           & 78k           & 1          & Fashion           & single-label  & 121 \\
UPMC Food-101 \cite{upmcFood101} &100k           & 100k          & 1          & Food              & single-label  & 101 \\
SNLI-VE \cite{Xie2019VisualEA}   & 30k            & 565k         & 1          & General           & single-label  & 3 \\
\hline
\end{tabular}
\vspace*{-5mm}
\end{table}

Large fashion datasets with image and text features are summarized in \autoref{tab:fashion}.
To the best of our knowledge, GLAMI-1M is the largest image-text dataset in terms of items and the most diverse dataset in terms of languages.
GLAMI-1M also offers the highest number of categories (191) for classification.
The only other multilingual fashion image-text dataset, Fashion-MMT \cite{fashionMMT}, is bilingual and ten times smaller in the number of items. \\

Other Fashion datasets without text annotations include:
DeepFashion2 \cite{deepfashion2} contains 800k diverse photos with clothing segmentation metadata.
Clothing-1M \cite{clothing1m} contains 1M product images with majority noisy class (14) labels.
MVC \cite{mvc} dataset of 161k items for view-invariant clothing retrieval, classification (23), colors (13), attribute prediction.
ModaNet \cite{modanet} is a 55k image segmentation dataset.
Fashionpedia \cite{fashionpedia} is a 45k image dataset with fine-grained apparel attribute (294) prediction, segmentation (27 categories, 19 parts), and an ontology.
StreetStyle \cite{streetStyle} is a 45k image dataset with various attributes including category (7).
DeepFashion3D \cite{deepfashion3d} is a 2k image to 3D reconstruction dataset with annotations including 10 categories.
Colorful-Fashion \cite{Liu2014FashionPW} is 2k image dataset for segmentation into 23 categories, 13 colors.

\begin{table}[h!]
\centering
\small
    \setlength{\tabcolsep}{2pt}
\renewcommand{\arraystretch}{1.2}
\caption{\label{tab:fashion} Overview of publicly available fashion product datasets with image and text features. GLAMI-1M is the biggest, most fine-grained, and uniquely multilingual fashion dataset.}
\vspace{1mm}
\begin{tabular}{|m{3.3cm}|c|c|m{6.4cm}|c|}
\hline
Dataset & Items & Imgs & Features & Langs\\
\hline
GLAMI-1M             & 1.11M & 968k &  image, name, description, class (191) & 13 \\
FACAD \cite{facad}             & 130k & 993K  & image, description, class (78) & 1 \\
Fashion-MMT \cite{fashionMMT}  & 110k & 853k & image, description with noisy translations, class (78), attributes & 2  \\
Fashion550k \cite{fashion550k} & 550k & 408k & image (in-the-wild), user comments, garment class, attributes, other metadata & 1  \\
Neti-look \cite{netilook}      & 350k  & 355k &  image (in-the-wild), comments & 1 \\
FashionGen \cite{fashiongen}   & 78k & 325k  & image, description, class (121) & 1 \\
Amazon Fashion Products 2020 \cite{amazonFashionProducts2020} & 132k & 132k+  & multiple images, name, other & 1 \\
Fashion IQ \cite{fashionIQ}    & 50k  & 50k & image, description, attributes, relative caption & 1 \\
Fashion Product Images \cite{fashionProductImages} & 44k & 44k & image, name, description, class, other & 1 \\
\hline
\end{tabular}
\end{table}

\begin{table}[htb]
\centering
\small
\setlength{\tabcolsep}{3pt}
\renewcommand{\arraystretch}{1.5}
\caption{\label{tab:columns} GLAMI-1M column descriptions, and unique value count in training and test sets.}
\vspace{1mm}
\begin{tabular}{ | m{1.9cm} | m{8.1cm} | r | r| }
\hline
          Name &                                                                                                                                                                                                                                                                           Description &  \# Train. &  \# Test \\
\hline
       item\_id &    Item integer identifier (Not unique). &   992528 &  116004 \\
      image\_id &       Image integer identifier. Products with duplicate images exists across different geos. &   882846 &   85577 \\
           geo &        Country code in lower case. It is a strong indicator of language used in the text. &       13 &      13 \\
          name &          Product name text. Often contains product's brand. &   752092 &  99783 \\
   description &        Product description text. It describes the product and advertises the product. &   656067 &   90313 \\
      category &        Integer category id label. &      191 &     191 \\
 category\_name &       Human readable category name label. &      191 &     191 \\
  label\_source &  Source of the class labels indicating label quality:\newline \textit{admin}, \textit{quality-check}, \textit{custom-tag}: human labels\newline \textit{combined-tag}, \textit{NaN}: machine labels -- simple rule based systems&        5 &       3 \\
\hline
\end{tabular}
\end{table}

\begin{figure}
\centering
\includegraphics[width=\textwidth,trim={0 2mm 0 0},clip]{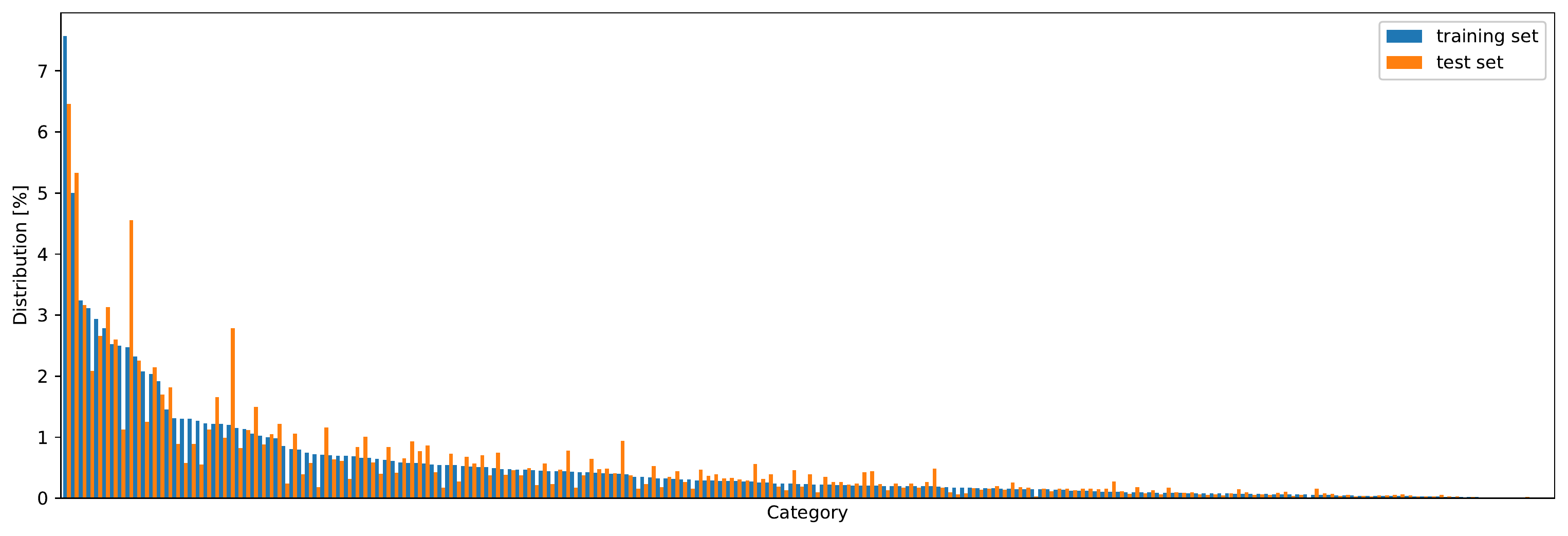}
\vspace*{-6mm}
\caption{Distribution of samples per category.
The distribution is mostly exponential, but steeper along the edges, so we regard this as a long tailed distribution.
} \label{fig:category}
\end{figure}

\section{Dataset Description}
\label{sec:GLAMI-1M}

We introduce GLAMI-1M: a 13-lingual image-text classification dataset of 1.10M items representing a product and its leaf category within GLAMI production catalog category tree.
Each item represents a product listing with:
 image, texts (name and description) in one of the 13 languages, category label and its label source.
Examples from the dataset are in \autoref{fig:exampleFront} and in the supplementary material.

\begin{table}[thb]
\centering
\caption{The 10 most and 10 least represented from the 191 total training set categories.}
\vspace{1mm}
\label{tab:top10cats}
        \setlength{\tabcolsep}{2.2pt}
\renewcommand{\arraystretch}{1.2}

\begin{tabular}{|l|r|r|}
\hline
                      Category name &  \# Train. &  \# Test \\
\hline
        mens-t-shirts-and-tank-tops &    75724 &    7497 \\
 womens-tops-tank-tops-and-t-shirts &    50000 &    6187 \\
                      mens-sneakers &    32385 &    3668 \\
                    womens-sneakers &    31137 &    2417 \\
                            dresses &    29350 &    3084 \\
                      baby-clothing &    27896 &    3631 \\
          womens-blouses-and-shirts &    25292 &    3017 \\
                       womens-pants &    24998 &    1305 \\
                            bikinis &    24712 &    5286 \\
                  womens-flip-flops &    23219 &    2612 \\
\hline
\end{tabular}
\begin{tabular}{|l|r|r|}
\hline
      Category name &  \# Train. &  \# Test \\
\hline
    mens-bath-robes &      211 &      26 \\
 mens-handkerchiefs &      200 &      11 \\
    mens-shoe-laces &      187 &       3 \\
     mens-umbrellas &      179 &      10 \\
    mens-suspenders &      171 &      19 \\
           broaches &      155 &      17 \\
        mens-chains &      122 &      16 \\
  mens-rubber-boots &       99 &      24 \\
      mens-earrings &       88 &      12 \\
     boys-tank-tops &       81 &      14 \\
\hline
\end{tabular}
\end{table}

\begin{figure}[thb]
\includegraphics[width=\textwidth]{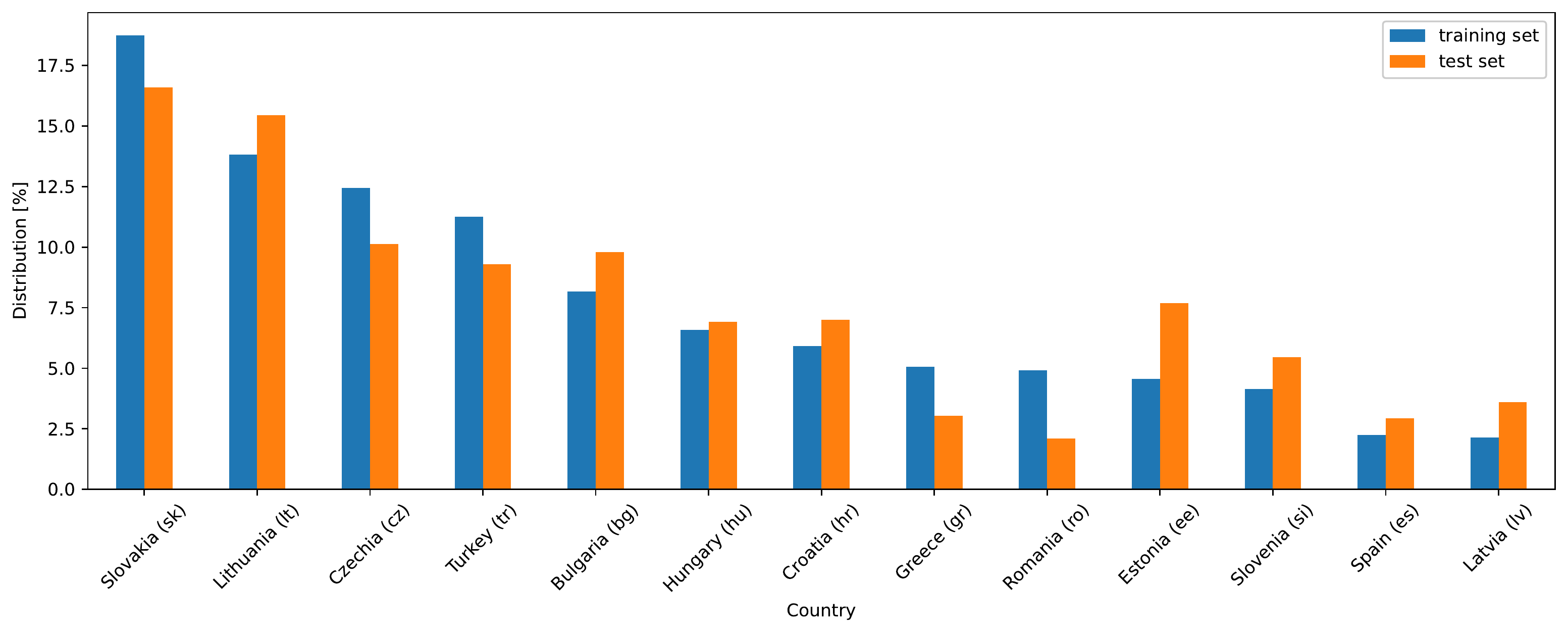}
\caption{
Sample distribution per country (geo), which roughly approximates the language distribution.
} \label{fig:geo}
\end{figure}

Items for the dataset were selected from the GLAMI catalogue in two phases: first, we sampled items with higher-quality human annotations (i.e. based on the label source). 100k of these items were randomly sampled for the test set.
Then items with labels from less reliable rule-based (heuristic) labeling systems were sampled proportionally to the catalog category distribution, in order to get a training set of 1M items. Zero overlap between the training and test set images and texts was checked via MD5 hashes and cosine similarity threshold of CLIP embeddings \cite{clip,mclip}. See the source code and the supplementary material for details. Text was preprocessed by removing backslashes, braces, brackets, semicolons, angle brackets, and replacing line ends, carriage returns and forward slashes with a space.

\autoref{tab:columns} describes the dataset's data columns with the numbers of unique values. The training set may contain several records describing the same item (i.e. records with the same \textit{item\_id}) -- e.g. because unisex items appear in both men's and women's category variants.
The test set contains only consistent human-label annotations without such duplicate records (with same \textit{item\_id}). However, up to tens of items still have the same \textit{image\_id}, since the same products are sometimes sold by multiple e-shops within the same or different country. In these cases the items have a different \textit{item\_id}. The classes are fine-grained: 15 categories of women shoes and total 191 categories in contrast to FashionGen's 121. The class distribution is long tailed, as shown in \autoref{fig:category}. The 10 most and 10 least frequent training set categories can be found in \autoref{tab:top10cats}. \autoref{fig:geo} shows a train-test distribution shift in number of samples per country. The distribution of product name and description lengths is illustrated in  \autoref{fig:nameLength}.  For the distribution of label source, please see the supplementary material.  

The dataset is primarily shared in a compact 10GB archive with 228x298px images in JPEG format. Larger 800x800px resolution variants are available in a separate archive.


\begin{figure}[thb]
\centering
\includegraphics[width=\textwidth]{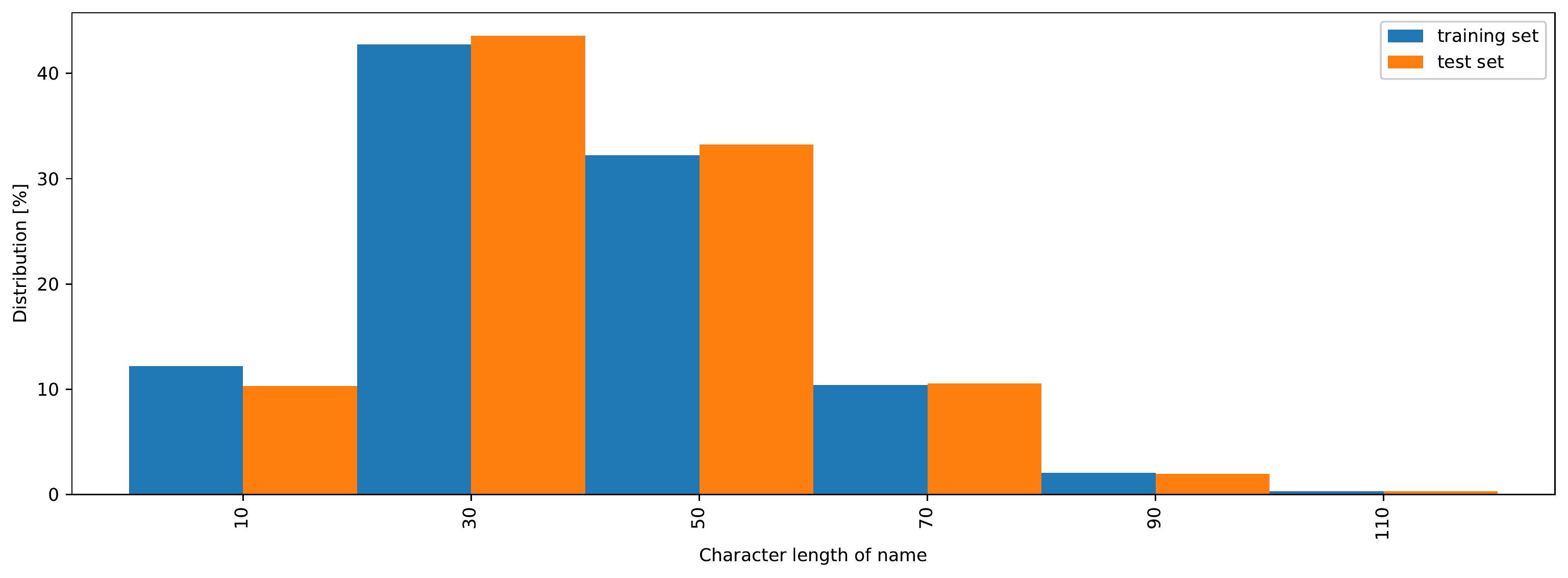}\\
\includegraphics[width=\textwidth,trim={0 2.5mm 0 0},clip]{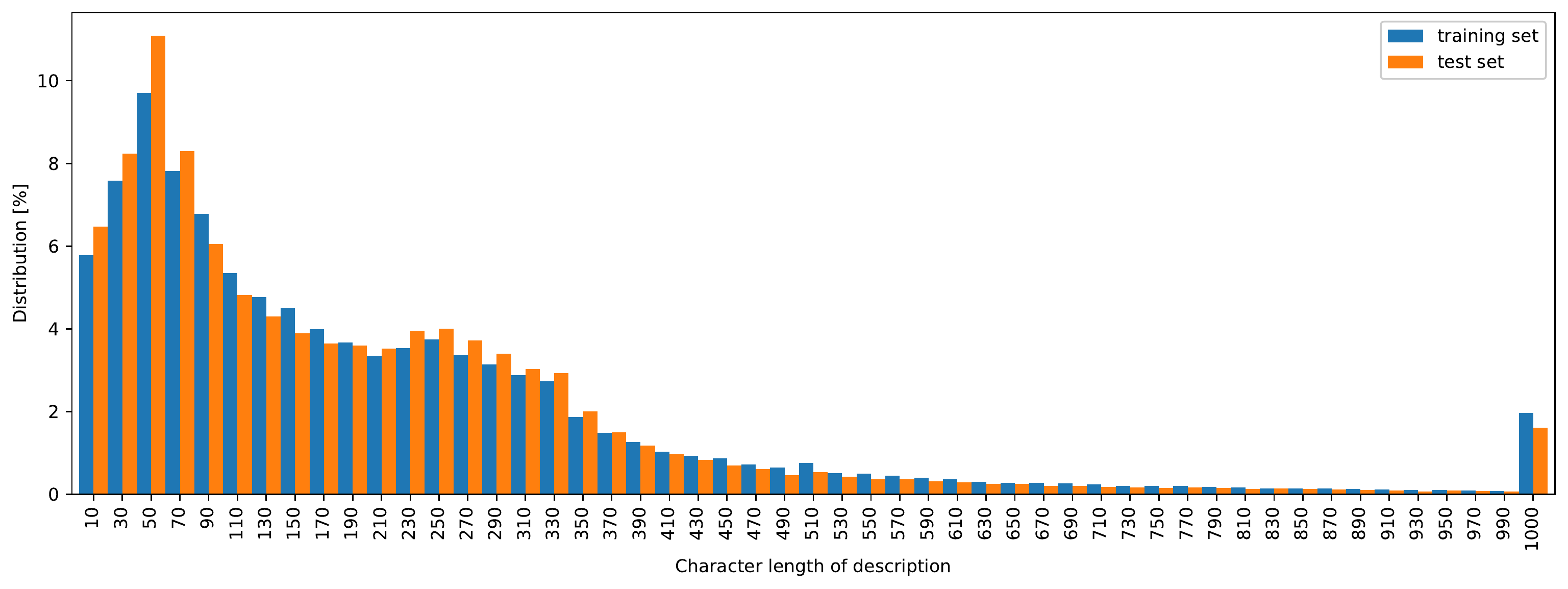}
\vspace*{-2mm}
\caption{Distribution of \textit{name} (top) and \textit{description} (bottom) length in characters in the training and test set. For \textit{name} the median is 38 characters, 
for \textit{description} the median is 150. Note that the last bin for description contains all the samples longer than 990 characters (up to 4000).}
\label{fig:nameLength}
\end{figure}


Together with the dataset, we set up a \textbf{public benchmark}\footnote{\label{note:leaderboard}accessible from \href{https://github.com/glami/glami-1m}{the repository}.} for multilingual image-text  classification. The benchmark's primary score is the test set accuracy. The benchmark allows using pre-trained models and additional training data, if explicitly stated in the method description. Initial results for baseline classification methods are provided in \autoref{sec:experiments_classification}.

Additionally, we provide baseline generative models for text-conditioned image generation, as described in \autoref{sec:experiments_imagen}, and baseline models and results for machine translation in the supplementary material.

\section{Experiments}
\label{sec:experiments}


\subsection{Multimodal Classification}
\label{sec:experiments_classification}
Classification is one the fundamental tasks of supervised learning \cite{sen2020supervised}. Multimodal classification models process inputs of several different modalities. In our benchmark the inputs come from three \emph{modalities}: textual (title + description), visual (image) and categorical (label source). The label source could be used as a meta information for training methods like sample weighting \cite{metaWeightNetLA} or label correction \cite{metaLabelCorrection}, however these experiments are beyond the scope of this paper. For baseline we have chosen EmbraceNet \cite{embracenet}, a robust model essentially capable of taking encoded inputs from any modality and automatically combining them into a single model. In all experiments, the model was trained for two epochs (early stopping) with the Adam optimizer and the internal EmbraceNet dimension set to 512. 

For the encoding of various modalities we have relied on well tested, publicly available, pre-trained models. We have encoded the textual inputs with the \emph{small} variant of the mT5 model \cite{xue-etal-2021-mt5}, which has been pretrained on a superset of the languages in our dataset. We encoded with maximum length of 32 tokens, which resulted in $(32 \times 512 = 16384)$ dimensional embeddings of the concatenated title + description. For the image inputs we have used a pretrained \emph{ResNeXt-50 32x4d} model \cite{resnext}, which after the last max pooling layer gives $2048$ dimensional embeddings. We finetuned ResNext, but froze mT5.

To better understand the quality of the input features, we have trained several versions of EmbraceNet by dropping one or multiple modalities from the input and by training on human-labels only or including the noisy labels too, see \autoref{tab:embracenets}. The best top-1 accuracy of 0.697 was achieved with the combination of both text and image and by including the noisy labels, while separately the image features outperformed the textual inputs. We note that we did not tune the probabilities of the fusion process in EmbraceNet \cite{embracenet}. The probability of the docking layers of each modality being included was thus $1/2$ for the bi-modal version. 
To see how EmbraceNet trained on images compares to the ResNeXt-50 32x4d model, we used a pre-trained ResNext and finetuned it on our dataset. Since in this case EmbraceNet essentially replaces the last fully connected layer in ResNext with several layers, thus increasing the number of parameters, the image-only version of EmbraceNet outperformed the original architecture. The presence of the noisy labels only has a small impact on the performance of EmbraceNet.

 A weak zero-shot CLIP baseline is available in the supplementary material.

\begin{table}
\centering
\caption{Top-k accuracies of EmbraceNet with various input modalities, trained either on all labels (\textit{all}) or human-labeled samples only (\textit{hum.}).}
\vspace{1mm}
\renewcommand{\arraystretch}{1.1}
\label{tab:embracenets}
\begin{tabular}{lcccc}
\hline
Included modality/model & Top-1 (all) &   Top-5 (all)  & Top-1 (hum.) &   Top-5 (hum.) \\
\hline
Text + Image & \textbf{0.697} & 0.940 & 0.694 & 0.932  \\
Image & 0.685 & \textbf{ 0.948} & 0.679 & 0.943  \\
Text &  0.593 & 0.840 &  0.613 & 0.849  \\
Finetuned ResNeXt-50 32x4d & 0.631 &  0.935 & 0.642 & 0.933 \\
\hline
\end{tabular}
\end{table}

\subsection{Text-Conditional Image Generation}
\label{sec:experiments_imagen}
The area of image generation conditioned on text has recently attracted much attention \cite{imagen,dall-e-2,rombach2022high}. Our dataset can be used for this task. We have trained a "small" version of the Imagen-like model \cite{imagen} on a single NVIDIA T4 GPU over 72 hours on 884k images and 992k texts. This underscores the position of our dataset in the matter of its size. It lies on the border between extremely large datasets, allowing to push the envelope of the state-of-the-art in machine learning,  and datasets compact enough to train a model on a single GPU in days. 

\begin{figure}[thb]
    \centering
    \caption{Cherry-picked images generated by the Imagen-like model 
    with the corresponding country codes, 500 time steps of diffusion. Large images are the generated ones, the two smaller are the closest images from train set based on \emph{ResNeXt-50 32x4d} embeddings.}
    \label{tab:imagen}
    \vspace{1mm}
    \setlength{\tabcolsep}{1pt}
\begin{tabular}{cccccccccc}
\hline
 bg & bg & bg & bg & cz & cz & es & es & hr & hu \\
 \includegraphics[width=1.2cm]{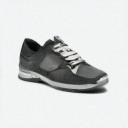}                                                               & \includegraphics[width=1.2cm]{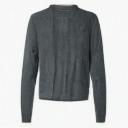}                                                               & \includegraphics[width=1.2cm]{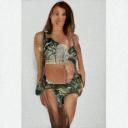}                                                               & \includegraphics[width=1.2cm]{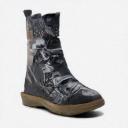}                                                               & \includegraphics[width=1.2cm]{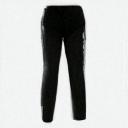}                             
 &
 \includegraphics[width=1.2cm]{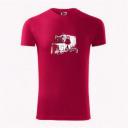}                                                              & \includegraphics[width=1.2cm]{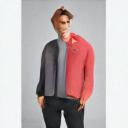}                                                               & \includegraphics[width=1.2cm]{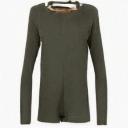}                                                               & \includegraphics[width=1.2cm]{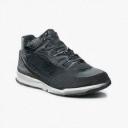}                                                               & \includegraphics[width=1.2cm]{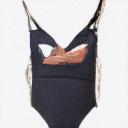}                              
 \\
 \includegraphics[width=0.6cm]{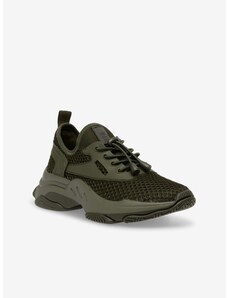}\includegraphics[width=0.6cm]{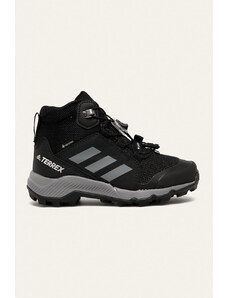} & \includegraphics[width=0.6cm]{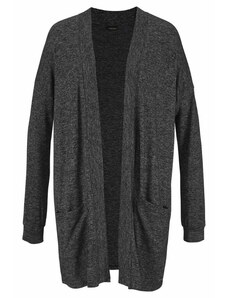}\includegraphics[width=0.6cm]{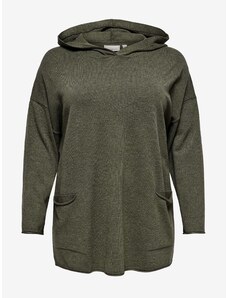} & \includegraphics[width=0.6cm]{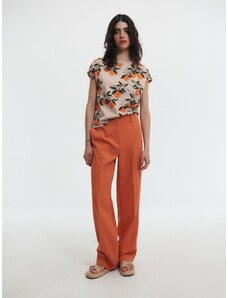}\includegraphics[width=0.6cm]{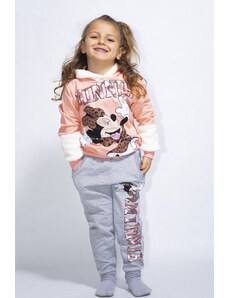} & \includegraphics[width=0.6cm]{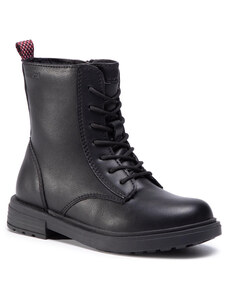}\includegraphics[width=0.6cm]{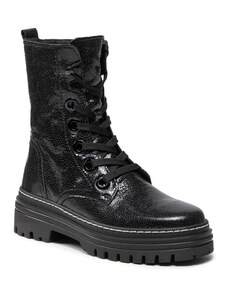} & \includegraphics[width=0.6cm]{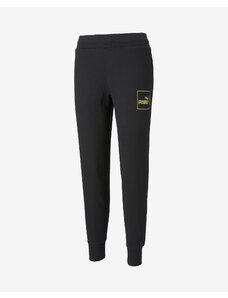}\includegraphics[width=0.6cm]{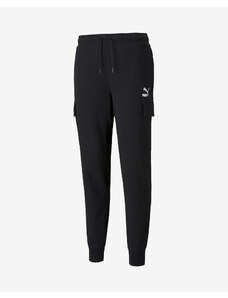} &
 \includegraphics[width=0.6cm]{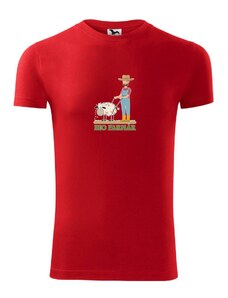}\includegraphics[width=0.6cm]{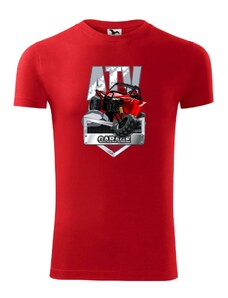} & \includegraphics[width=0.6cm]{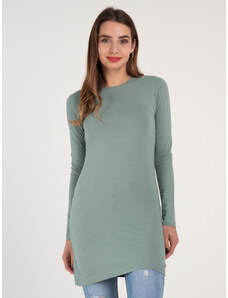}\includegraphics[width=0.6cm]{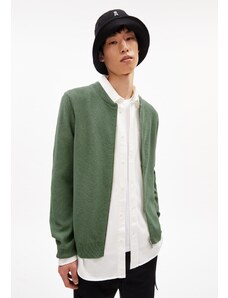} & \includegraphics[width=0.6cm]{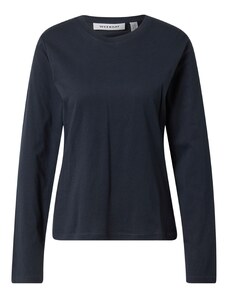}\includegraphics[width=0.6cm]{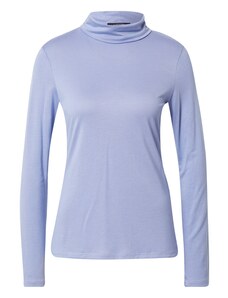} & \includegraphics[width=0.6cm]{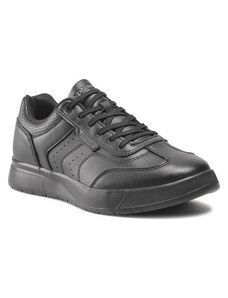}\includegraphics[width=0.6cm]{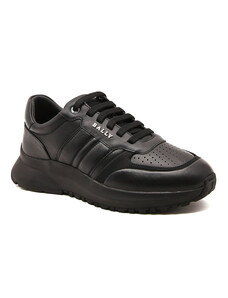} & \includegraphics[width=0.6cm]{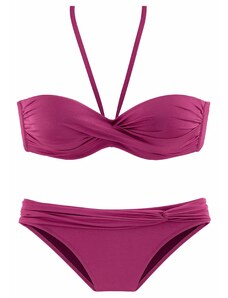}\includegraphics[width=0.6cm]{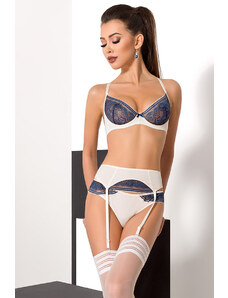} \\
 lt & lt & lt & lt & ro & si & sk & sk & sk & tr                                                                                                                   \\
 \includegraphics[width=1.2cm]{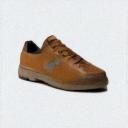}                                                               & \includegraphics[width=1.2cm]{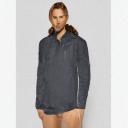}                                                               & \includegraphics[width=1.2cm]{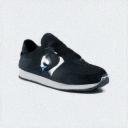}                                                               & \includegraphics[width=1.2cm]{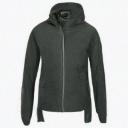}                                                               & \includegraphics[width=1.2cm]{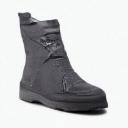}                                                              &
 \includegraphics[width=1.2cm]{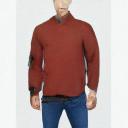}                                                               & \includegraphics[width=1.2cm]{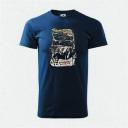}                                                               & \includegraphics[width=1.2cm]{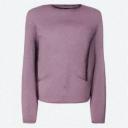}                                                              & \includegraphics[width=1.2cm]{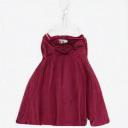}                                                               & \includegraphics[width=1.2cm]{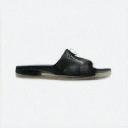}                                                              \\
 \includegraphics[width=0.6cm]{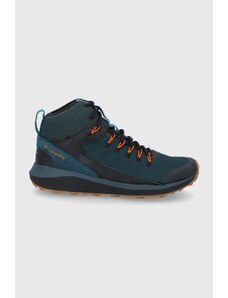}\includegraphics[width=0.6cm]{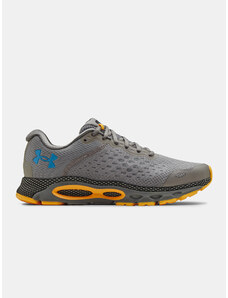} & \includegraphics[width=0.6cm]{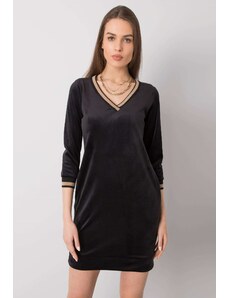}\includegraphics[width=0.6cm]{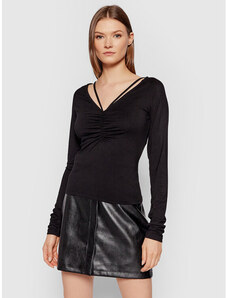} & \includegraphics[width=0.6cm]{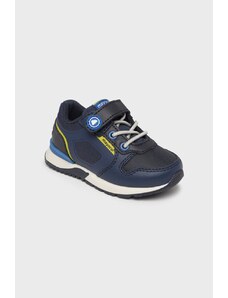}\includegraphics[width=0.6cm]{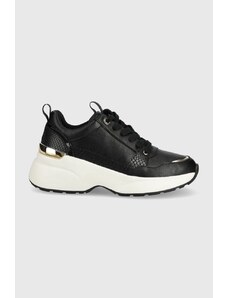} & \includegraphics[width=0.6cm]{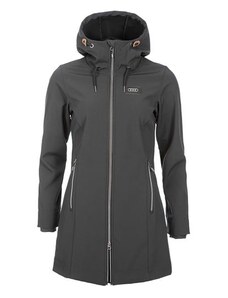}\includegraphics[width=0.6cm]{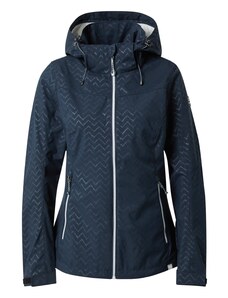} & \includegraphics[width=0.6cm]{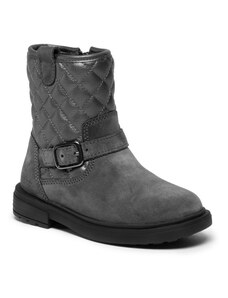}\includegraphics[width=0.6cm]{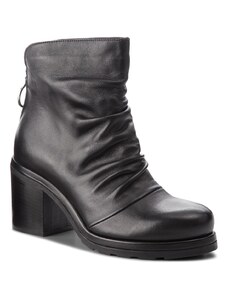} &
 \includegraphics[width=0.6cm]{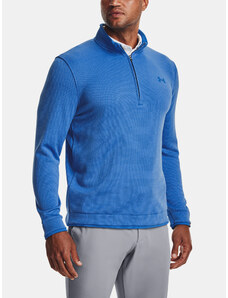}\includegraphics[width=0.6cm]{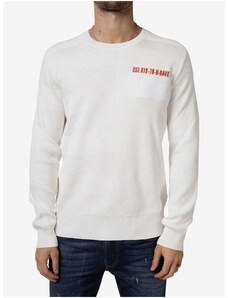} & \includegraphics[width=0.6cm]{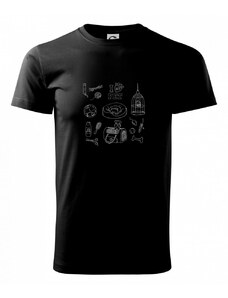}\includegraphics[width=0.6cm]{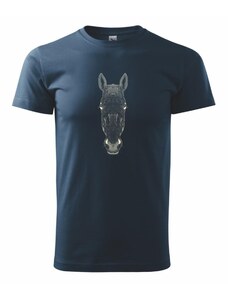} & \includegraphics[width=0.6cm]{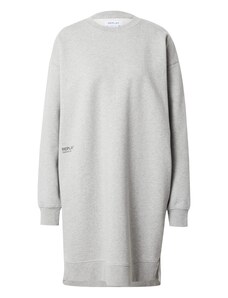}\includegraphics[width=0.6cm]{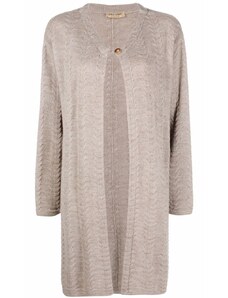} & \includegraphics[width=0.6cm]{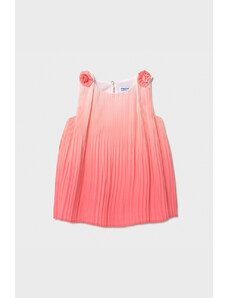}\includegraphics[width=0.6cm]{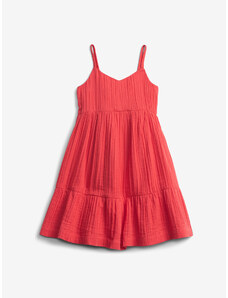} & \includegraphics[width=0.6cm]{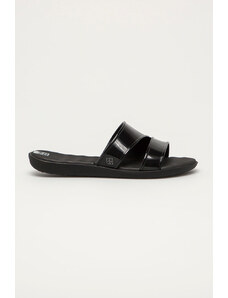}\includegraphics[width=0.6cm]{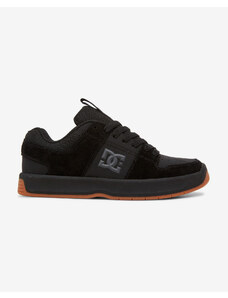} \\
\hline
\end{tabular}

\end{figure}

\begin{figure}[thb]
    \centering
    \caption{Random samples of images generated by the Imagen-like model 
    with the corresponding country codes, 500 time steps of diffusion.}
    \label{tab:imagen-random}
    \vspace{1mm}
    \setlength{\tabcolsep}{1pt}
\begin{tabular}{cccccccccc}
\hline

sk &     tr &     sk &     bg &     cz &     si &     cz &     tr &     bg &     ee \\

     \includegraphics[width=1.2cm]{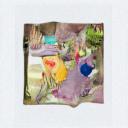} & \includegraphics[width=1.2cm]{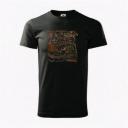} & \includegraphics[width=1.2cm]{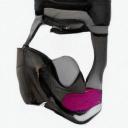} & \includegraphics[width=1.2cm]{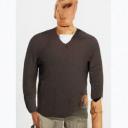}  & \includegraphics[width=1.2cm]{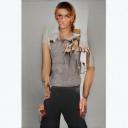} &
     \includegraphics[width=1.2cm]{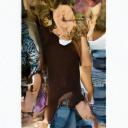}  & \includegraphics[width=1.2cm]{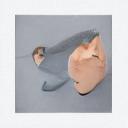} & \includegraphics[width=1.2cm]{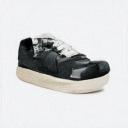} & \includegraphics[width=1.2cm]{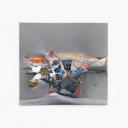}  & \includegraphics[width=1.2cm]{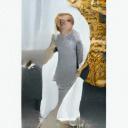}   \\
     
sk &     es &     sk &     sk &     sk &     sk &     cz &     hu &     si &     bg \\
     \includegraphics[width=1.2cm]{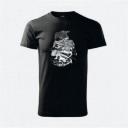}  & \includegraphics[width=1.2cm]{imagen/es-6155059_1.jpg}  & \includegraphics[width=1.2cm]{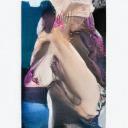} & \includegraphics[width=1.2cm]{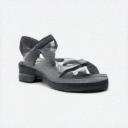} & \includegraphics[width=1.2cm]{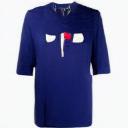} &
     \includegraphics[width=1.2cm]{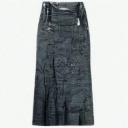}  & \includegraphics[width=1.2cm]{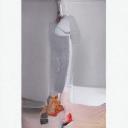} & \includegraphics[width=1.2cm]{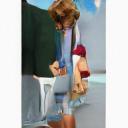} & \includegraphics[width=1.2cm]{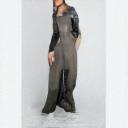}   & \includegraphics[width=1.2cm]{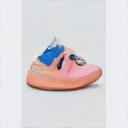}  \\

\hline
\end{tabular}

\end{figure}

\begin{figure}
    \centering
    \caption{Images generated by the Imagen-like model for the input "sneakers" translated into all 13 languages, 500 time steps of diffusion.}
    \vspace{1mm}
    \label{tab:imagen_snakers}
    \scriptsize
        \setlength{\tabcolsep}{1pt}
\renewcommand{\arraystretch}{1}
    \begin{tabular}{lllllll}
    \hline
     bg: \raisebox{-0.07cm}{\includegraphics[height=0.2cm]{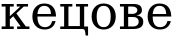}}                                   & cz: tenisky                                    & ee: tossud                                     & es: las zapatillas                             & gr: \raisebox{-0.07cm}{\includegraphics[height=0.2cm]{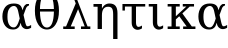}}
     & hr: tenisice                                   & hu: tornacipő\\
     \includegraphics[width=1.8cm]{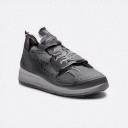} & \includegraphics[width=1.8cm]{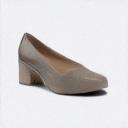} & \includegraphics[width=1.8cm]{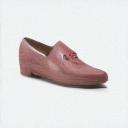} & \includegraphics[width=1.8cm]{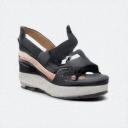} & \includegraphics[width=1.8cm]{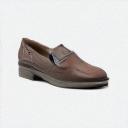} & 
     \includegraphics[width=1.8cm]{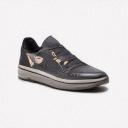} & \includegraphics[width=1.8cm]{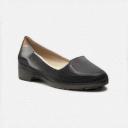} 
     \\
    lt: sportbačiai                                & lv: kedas                                      & ro: adidași & si: superge                                    & sk: tenisky                                    & tr: spor ayakkabı                              &                                   \\
     \includegraphics[width=1.8cm]{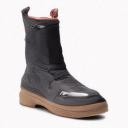} & \includegraphics[width=1.8cm]{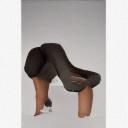} & \includegraphics[width=1.8cm]{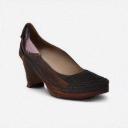} &
     \includegraphics[width=1.8cm]{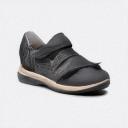} & \includegraphics[width=1.8cm]{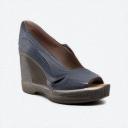} & \includegraphics[width=1.8cm]{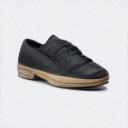} &                                           \\
    \hline
    \end{tabular}
\vspace{-4mm}
\end{figure}

We have trained a small cascading Denoising Diffusion Probabilistic Model \cite{denoisingDP} conditioned on text embeddings with two Unet models inside, each with the internal dimension of 128. We used the same text embeddings as in \autoref{sec:experiments_classification}. The generation happens in two steps, we first generate a 64x64 image from which we upscale to 128x128 pixels. We make this model publicly available, including its weights. See the \href{https://github.com/glami/glami-1m}{the source code} for more details.

We report a sample of visual results: \autoref{tab:imagen} shows images sampled on the texts from the test set. Another interesting property of the generator is the novelty of the generated images. We have looked up the two closest images using visual embeddings in the training set for each of the generated images and we were unable to find identical looking images in the train set. Let us underscore that not only the generated images are not pixel perfect replicas of the train samples, but they are quite far from the training samples even by human standards - not just L2 distance caused by imperceptible noise. We have cherry-picked the samples in this table to show that the model learned to draw product images that appear almost realistic. For an unbiased sample of images generated from the test set, see more examples in the \autoref{tab:imagen-random}.
About one third of the images generated by the model appears realistic based on a sample of about 1000 images checked by hand. Furthermore, we have experimented with the text-conditioning and generated images for various phrases translated into all of the languages in the dataset. In about one third of the texts the conditioning failed, in about a third it reflected the correct piece of clothing, but the style was wrong. In other words, high-level category such as "shoes" was correct, but a low-level one such as "sneakers" was often wrong. In about the last third of cases it worked to obtain a realistic sample, see for example \autoref{tab:imagen_snakers}. 

\section{Conclusion}

The paper introduced GLAMI-1M: the largest publicly available multilingual image-text classification dataset and the largest image-text dataset in the fashion domain. 
Thanks to its characteristics, the dataset has the potential to accelerate research in several areas of machine learning, including multilingual image-text classification, text-conditional image generation and multilingual machine translation. For example, it can be used as an alternative to Recipe1M+ \cite{recipe1m+} adding the aspect of multilinguality, or as a larger alternative to FashionGen \cite{fashiongen} and other datasets in the fashion domain.

Experiments on multimodal image classification in \autoref{sec:experiments_classification} show the dataset presents a challenging problem. Together with the dataset, we introduce a benchmark with baseline results and pre-trained models available, and we invite everyone to evaluate their models in the public leaderboard. 

Additional experiments on text-conditional image generation and multi-language machine translation are described in \autoref{sec:experiments_imagen}  and in the supplementary material respectively. The experiments illustrate the dataset's usefulness for other tasks than classification. Pre-trained models and code for the tasks are also shared with the paper.

Other relevant problems left for future work include long-tail learning, adaptation to prior shift, learning from a combination of trusted (human) and noisy (rule-based) annotations.

\newpage

\end{document}


\newgeometry{twoside,headsep=3mm,papersize={410pt,620pt},inner=15mm,outer=6mm,top=3mm,includehead,bottom=1mm,heightrounded}
\maketitle

\section{Supplementary Tables and Figures}

\autoref{tab:examples} visualizes dataset examples.
\autoref{tab:labelSources} shows distribution of the \textit{label\_source} column.
\autoref{tab:clip} results from zero-shot CLIP model baseline.
\autoref{tab:accuracy_per_geo} results of EmbraceNet with image and text on the input, stratified per-language.
\autoref{tab:countryCodes} lists country names corresponding to country codes for all 13 dataset languages.

\section{Machine Translation Experiments}

Some of the samples appear in the dataset in multiple languages. In these cases the underlying product is the same, but the title and description have been written in a different language often by a different seller. These samples can be identified by the image\_id column. They have a different item\_id, but identical image\_id. The publicly available datasets for machine translation are typically based on Wikipedia, news feeds or scraped websites  and are scarcely available for non-English languages \cite{wmt,tatoeba,zhang2020improving,lison2016opensubtitles2016,michel2018mtnt}. The pair-wise counts of samples in our training and test splits can be found in \autoref{tab:mcounts_train} and \autoref{tab:mcounts_test} respectively. For some of the language pairs the counts are orders of magnitude higher than what was previously available.

\restoregeometry

\begin{sidewaystable}[h]
\centering
\scriptsize
\caption{\label{tab:examples} Examples from GLAMI-1M.}
\begin{tabular}{|m{0.6cm}|m{1cm}|m{0.3cm}|m{2cm}|m{2.2cm}|m{1cm}|l|l|}
\hline
 item\_id &  image\_id & geo &   name &    description &  category &     category\_name & label\_source \\
\hline
  517876 &    488425 \includegraphics[width=1.0cm]{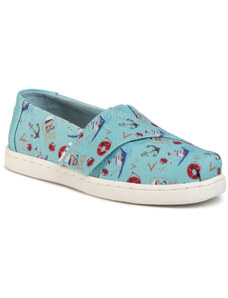} &  gr &                             Κλειστά παπούτσια TOMS &  Κλειστά παπούτσια TOMSΚλειστά παπούτσια TOMS -... &      2811 &                          boys-shoes &          NaN \\
\hline
  989034 &    863506 \includegraphics[width=1.0cm]{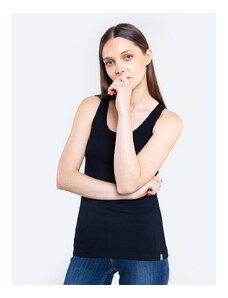} &  lt &  Big Star Woman's Singlet T-shirt 150048 Knitte... &  Material: 95\%COTTON5\%ELASTANE Washing instruct... &     53403 &  womens-tops-tank-tops-and-t-shirts &        admin \\
\hline
  483208 &    455633 \includegraphics[width=1.0cm]{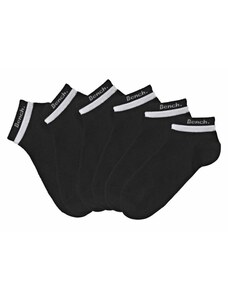} &  gr &                        BENCH Κάλτσες μαύρο   λευκό &  Υλικό: Ζέρσεϊ Έξτρα: Κεντημένο λογότυπο, Μαλακ... &       132 &                        womens-socks &        admin \\
\hline
 1009868 &    876723 \includegraphics[width=1.0cm]{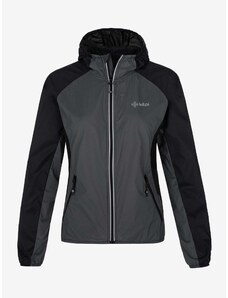} &  si &             Kilpi Ženske športne jakne črna Rosa-W &                                                 &     86531 &                womens-sport-jackets &   custom-tag \\
 \hline
  586781 &    544307 \includegraphics[width=1.0cm]{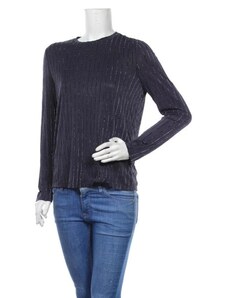} &  hu &                                      Női blúz ONLY &                                Új termék címkével. &         6 &           womens-blouses-and-shirts &          NaN \\
  \hline
  1121212 &    951403 \includegraphics[width=1.0cm]{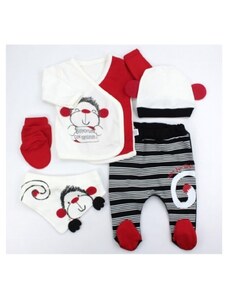} &  tr &             Nonna Baby Cute Monnet 5 Li Zıbın Seti &  Yeni sezon 5 parça zıbın seti,0-3 ay \%100 pamu... &     39412 &                       baby-clothing &   custom-tag \\

\hline
\end{tabular}
\end{sidewaystable}

\begin{table}[h]
\centering
\caption{\label{tab:labelSources} Distribution of values in \textit{label\_source} column.}
\begin{tabular}{lrr}
\hline
  label\_source & training set [\%] & test set [\%] \\
\hline
     custom-tag &          40.8 &      54.3 \\
         admin &          34.3 &      45.7 \\
           NaN &          24.4 &       0.0 \\
  combined-tag &           0.5 &       0.0 \\
 quality-check &           0.1 &       0.1 \\
\hline
\end{tabular}
\end{table}

\begin{table}
\centering
\caption{Top-k accuracies of CLIP zero-shot classification baseline with various input modalities. Image+text variant is classification using unnormalized embedding vector summation of CLIP image and text embeddings.  We used prompts "A photo of a {category}, a type of fashion product" as targets. We used aligned image (ViT-B/32) \cite{clip} and multilingual text (XLM-Roberta-Large-Vit-B-32) \cite{mclip} CLIP embeddings.}
\vspace{1mm}
\renewcommand{\arraystretch}{1.1}
\label{tab:clip}
\begin{tabular}{lcccc}
\hline
Included modality/model & Top-1 &   Top-5 \\
\hline
Text + Image & \textbf{0.323} & \textbf{0.745} \\
Image & 0.289 & 0.718  \\
Text &  0.265 & 0.585  \\
\hline
\end{tabular}
\end{table}

\begin{table}[h]
    \centering
    \tiny
    \caption{Top-k accuracies of EmbraceNet with text and image inputs, trained on all labels, stratified per-language. We observe maximal 6\% difference in Top-5 accuracy across different countries and 21\% in Top-1 accuracy. We speculate that the reason may be variable quality of text embeddings and different distributions of test set samples between the countries.}
    \label{tab:accuracy_per_geo}

\begin{tabular}{lrrrrrrrrrrrrr}
\hline
  k  &   cz &   sk &   ro &   gr &   hu &   bg &   hr &   es &   lt &   si &   lv &   tr &   ee \\
\hline
 1 & 0.592 & 0.805 & 0.646 & 0.589 & 0.670 & 0.726 &  0.739 & 0.626 & 0.685 & 0.683 & 0.672 & 0.720 & 0.659  \\
 5 & 0.905 & 0.967 & 0.924 & 0.911 & 0.941 & 0.961 &  0.956 & 0.920 & 0.941 & 0.946 & 0.922 & 0.932 & 0.928  \\
\hline
\end{tabular}

\end{table}

\begin{table}[h]
    \centering
    \scriptsize
    \caption{Pairwise counts of samples in multiple languages, training set.}
    \label{tab:mcounts_train}
    
    \begin{tabular}{lrrrrrrrrrrrrr}
\hline
    &   cz &   sk &   ro &   gr &   hu &   bg &   hr &   es &   lt &   si &   lv &   tr &   ee \\
\hline
 cz &    0 & 4669 & 1249 &  977 & 2712 & 1797 & 1986 &  784 & 1877 & 1022 &  723 &   10 & 1148 \\
 sk & 4669 &    0 & 2231 &  635 & 3882 & 2366 & 2526 &  722 & 2557 & 1024 &  485 &   31 & 1485 \\
 ro & 1249 & 2231 &    0 &  766 & 1417 & 1276 & 1748 &  416 &  731 &  644 &  238 &  214 &  370 \\
 gr &  977 &  635 &  766 &    0 &  825 & 1901 & 1231 &  648 & 1317 &  444 &  234 &    5 &  461 \\
 hu & 2712 & 3882 & 1417 &  825 &    0 & 5458 & 2232 & 1076 & 3137 & 1937 &  613 &  730 & 1188 \\
 bg & 1797 & 2366 & 1276 & 1901 & 5458 &    0 & 3235 & 1416 & 6880 & 2983 & 1086 &  548 & 3297 \\
 hr & 1986 & 2526 & 1748 & 1231 & 2232 & 3235 &    0 & 1174 & 3319 & 3408 &  933 &    0 & 2095 \\
 es &  784 &  722 &  416 &  648 & 1076 & 1416 & 1174 &    0 &  860 &  381 &  710 &  322 & 1099 \\
 lt & 1877 & 2557 &  731 & 1317 & 3137 & 6880 & 3319 &  860 &    0 & 5640 & 5538 &   10 & 9176 \\
 si & 1022 & 1024 &  644 &  444 & 1937 & 2983 & 3408 &  381 & 5640 &    0 & 1968 &    9 & 1737 \\
 lv &  723 &  485 &  238 &  234 &  613 & 1086 &  933 &  710 & 5538 & 1968 &    0 &    5 & 5085 \\
 tr &   10 &   31 &  214 &    5 &  730 &  548 &    0 &  322 &   10 &    9 &    5 &    0 &    4 \\
 ee & 1148 & 1485 &  370 &  461 & 1188 & 3297 & 2095 & 1099 & 9176 & 1737 & 5085 &    4 &    0 \\
\hline
\end{tabular}

\end{table}

\begin{table}[h]
    \centering
    \scriptsize
    \caption{Pairwise counts of samples in multiple languages, test set.}
    \label{tab:mcounts_test}

\begin{tabular}{lrrrrrrrrrrrrr}
\hline
    &   cz &   sk &   ro &   gr &   hu &   bg &   hr &   es &   lt &   si &   lv &   tr &   ee \\
\hline
 cz &    0 & 4669 & 1249 &  977 & 2712 & 1797 & 1986 &  784 & 1877 & 1022 &  723 &   10 & 1148 \\
 sk & 4669 &    0 & 2231 &  635 & 3882 & 2366 & 2526 &  722 & 2557 & 1024 &  485 &   31 & 1485 \\
 ro & 1249 & 2231 &    0 &  766 & 1417 & 1276 & 1748 &  416 &  731 &  644 &  238 &  214 &  370 \\
 gr &  977 &  635 &  766 &    0 &  825 & 1901 & 1231 &  648 & 1317 &  444 &  234 &    5 &  461 \\
 hu & 2712 & 3882 & 1417 &  825 &    0 & 5458 & 2232 & 1076 & 3137 & 1937 &  613 &  730 & 1188 \\
 bg & 1797 & 2366 & 1276 & 1901 & 5458 &    0 & 3235 & 1416 & 6880 & 2983 & 1086 &  548 & 3297 \\
 hr & 1986 & 2526 & 1748 & 1231 & 2232 & 3235 &    0 & 1174 & 3319 & 3408 &  933 &    0 & 2095 \\
 es &  784 &  722 &  416 &  648 & 1076 & 1416 & 1174 &    0 &  860 &  381 &  710 &  322 & 1099 \\
 lt & 1877 & 2557 &  731 & 1317 & 3137 & 6880 & 3319 &  860 &    0 & 5640 & 5538 &   10 & 9176 \\
 si & 1022 & 1024 &  644 &  444 & 1937 & 2983 & 3408 &  381 & 5640 &    0 & 1968 &    9 & 1737 \\
 lv &  723 &  485 &  238 &  234 &  613 & 1086 &  933 &  710 & 5538 & 1968 &    0 &    5 & 5085 \\
 tr &   10 &   31 &  214 &    5 &  730 &  548 &    0 &  322 &   10 &    9 &    5 &    0 &    4 \\
 ee & 1148 & 1485 &  370 &  461 & 1188 & 3297 & 2095 & 1099 & 9176 & 1737 & 5085 &    4 &    0 \\
\hline
\end{tabular}

\end{table}

For the baseline we use publicly available, pretrained M2M100 model \cite{m2m100}. M2M100 has been pretrained on the task of machine translation on a superset of the languages in our dataset. We have evaluated the BLEU score of the M2M100 model on the descriptions, on all of the possible pairs of languages in the test split of the dataset \autoref{tab:bleu_test}. We report the overall BLEU score of 1.87\%. The highest BLEU of 9.96\% was achieved in translation from Hungarian to Greek. 

In \autoref{tab:translations} are samples of the M2M100 Translations. We show a successful translation from Hungarian to Greek in sample no. 1 and then we show a failure from Czech to Slovak, where the model shortened the text much below the threshold of 32 tokens. Interestingly, this is a mistake that we saw quite often, for some reason the decoded sequence from the model came out much shorter than both the input and targets, this definitely brought the BLEU score down. Another frequent mistake was a change in units of measure, for example we saw the model translate 25 mm as 2.5 mm. On the other hand the dataset frequently contains brand names and if the model did not change them, its BLEU score increased.

This experiment could be reformulated as a multilingual text generation conditioned on the image or the title of the product. However, such models are beyond the scope of this paper.

\begin{table}[h]
    \centering
    \scriptsize
    \caption{BLEU scores (in \%) of the M2M100 model on all pairs of languages, on left is the source language, on the top is the target language.}
    \label{tab:bleu_test}

\begin{tabular}{lrrrrrrrrrrrrr}
\hline
    &     cz &     sk &     ro &     gr &     hu &     bg &     hr &     es &     lt &     si &     lv &     tr &     ee \\
\hline
 cz & nan    &   7.30 &   0.63 &   0.79 &   2.26 &   0.91 &   1.38 &   0.45 &   2.36 &   0.47 &   0.17 &   0.00 &   0.07 \\
 sk &   6.68 & nan    &   1.53 &   2.57 &   3.48 &   1.28 &   2.27 &   0.06 &   2.86 &   4.49 &   0.00 &   0.00 &   0.35 \\
 ro &   0.57 &   2.49 & nan    &   5.87 &   4.20 &   0.02 &   4.58 &   0.47 &   2.21 &   1.33 &   0.49 &   2.40 &   0.17 \\
 gr &   1.70 &   4.48 &   3.86 & nan    &   8.46 &   2.32 &   1.59 &   0.20 &   6.44 &   0.78 &   0.00 & nan    &   0.03 \\
 hu &   2.35 &   3.80 &   4.92 &   9.96 & nan    &   1.78 &   2.91 &   0.56 &   3.88 &   0.23 &   0.00 &   1.32 &   0.12 \\
 bg &   2.67 &   2.86 &   1.42 &   6.00 &   5.36 & nan    &   7.35 &   0.34 &   8.89 &   4.37 &   1.98 &   2.01 &   0.23 \\
 hr &   2.18 &   2.38 &   3.63 &   2.02 &   1.89 &   0.83 & nan    &   0.36 &   1.71 &   2.24 &   0.09 &   0.00 &   0.17 \\
 es &   0.11 &   0.07 &   0.27 &   0.45 &   0.35 &   0.08 &   0.22 & nan    &   0.00 &   0.16 &   0.26 &   1.64 &   0.03 \\
 lt &   2.56 &   3.64 &   2.29 &   3.93 &   3.30 &   4.23 &   2.13 &   0.00 & nan    &   0.34 &   0.01 &   0.00 &   0.05 \\
 si &   0.51 &   0.42 &   0.81 &   0.84 &   1.04 &   0.28 &   2.51 &   0.08 &   0.29 & nan    &   0.72 & nan    &   0.15 \\
 lv &   0.06 &   0.00 &   0.14 &   0.26 &   0.00 &   0.08 &   0.00 &   0.12 &   0.00 &   0.16 & nan    & nan    &   0.05 \\
 tr &   0.00 &   0.00 &   2.36 & nan    &   0.64 &   0.88 &   0.00 &   2.24 &   0.00 & nan    & nan    & nan    &   0.00 \\
 ee &   0.15 &   0.32 &   0.01 &   0.02 &   0.09 &   0.04 &   0.06 &   0.00 &   0.06 &   0.13 &   0.10 &   0.00 & nan    \\
\hline
\end{tabular}

\end{table}

\begin{table}[h]
    \centering
    \scriptsize
    \caption{Examples of M2M100 translations. Sample no. 1 is an example of a successful translation in the highest quality pair of languages based on the BLEU scores.}
    \label{tab:translations}

    \begin{tabular}{m{2.5cm} m{8cm}}
    \hline
    (Sample no.) Description & Text \\
    \hline
    (1) EN Translation & Color: white, Collection: Spring Summer 2020, Producer code: MM2T791, Fashion: Regular Fit \\
    \hline
    (1) HU Source Text & Szín: fehér , Kollekció: Tavaszi Nyár 2020 , Gyártókód: MM2T791, Fazon: Regular Fit \\ 
    \hline
    (1) GR Translation & Szín: λευκό, Kollekció: Tavaszi Nyár 2020, Gyártókód: MM2T791, Φά \\
    \hline
    (1) GR Target Text & Χρώμα: άσπρο, Συλλογή: Άνοιξη Καλοκαίρι 2020, Κωδικός παραγωγού: MM2T791, Μόδα: Regular Fit \\
    \hline
    \hline
    (2) EN Translation & A dress with an A-line skirt will liven up your look wherever you go. The delicate and understated look of this dress is completed by the gold zipper on the front. Thanks to its cut, it also conjures up a beautiful figure.\\
    \hline
    (2) CZ Source Text & Šaty s áčkovou sukní oživí Tvůj vzhled, ať půjdeš kamkoli. Jemný a decentní vzhled těchto šatů doplňuje zlatý zip na přední části. Díky svému střihu Ti navíc vykouzlí krásnou postavu.\\
    \hline
    (2) SK Translation & Šaty s áčkovou sukní oživí Tvůj vzhled, ať půjdeš kamkoli.\\
    \hline
    (2) SK Target Text & Šaty s áčkovou sukňou oživia Tvoj vzhľad, nech sa pohneš kamkoľvek. Jemný a decentný vzhľad týchto šiat dopĺňa zlatý zips na prednej časti. Vďaks svojmu strihu Ti navyše vyčarujú krásnu postavu. \\
    \hline
    \end{tabular}

\end{table}

\begin{table}
\centering
\caption{\label{tab:countryCodes} Country names corresponding to country codes}
\begin{tabular}{ll}
\hline
Country code (geo) & Country name \\
\hline
                cz &      Czechia \\
                sk &     Slovakia \\
                ro &      Romania \\
                gr &       Greece \\
                si &     Slovenia \\
                hu &      Hungary \\
                hr &      Croatia \\
                es &        Spain \\
                lt &    Lithuania \\
                lv &       Latvia \\
                tr &       Turkey \\
                ee &      Estonia \\
                bg &     Bulgaria \\
\hline
\end{tabular}
\end{table}

\newpage
\clearpage